\documentclass{tlp}

\usepackage{multirow}
\usepackage{array}

\newcommand{\soh}[3]{#3}

﻿\newcommand{\basicone}{base1}
\newcommand{\basictwo}{base2}
\usepackage{pifont}
\usepackage{graphicx}
\usepackage{lineno}
\usepackage{amsmath}
\usepackage{hyperref}
\usepackage{ascmac}
\usepackage{tikz}
\usetikzlibrary{arrows,shapes}
\usetikzlibrary{matrix}
\usetikzlibrary{positioning}
\usepackage{listings}
\newcommand{\ddreconf}{\textit{ddreconf}}
\newcommand{\paris}{\textit{PARIS}}

\newcommand{\clingo}{\textit{clingo}}
\newcommand{\recongo}{\textit{recongo}}

\newcommand{\code}[1]{\lstinline[basicstyle=\ttfamily]{#1}}

\newcommand{\OK}{\mbox{\textcolor{green}{\Pisymbol{pzd}{52}}}}
\newcommand{\KO}{\mbox{\textcolor{red}{\Pisymbol{pzd}{56}}}}

\begin{document}

\lefttitle{M.~Kato, T.~Schaub, T.~Soh, N.~Tamura, and M.~Banbara}

\jnlPage{1}{8}
\jnlDoiYr{2021}
\doival{10.1017/xxxxx}

\title[Dominating Set Reconfiguration with ASP]{Dominating Set Reconfiguration with Answer Set Programming}
    
\begin{authgrp}
\author{\gn{Masato} \gn{Kato}}
\affiliation{Nagoya University, Japan \email{kato.masato@nagoya-u.jp}}
\author{\gn{Torsten} \gn{Schaub}}
\affiliation{Universit{\"a}t Potsdam, Germany \email{torsten@cs.uni-potsdam.de}}
\author{\gn{Takehide} \gn{Soh}}
\affiliation{Kobe University, Japan \email{soh@lion.kobe-u.ac.jp}}
\author{\gn{Naoyuki} \gn{Tamura}}
\affiliation{Kobe University, Japan \email{tamura@kobe-u.ac.jp}}
\author{\gn{Mutsunori} \gn{Banbara}}
\affiliation{Nagoya University, Japan \email{banbara@nagoya-u.jp}}
\end{authgrp}

\history{\sub{xx xx xxxx;} \rev{xx xx xxxx;} \acc{xx xx xxxx}}

\maketitle

\begin{abstract}
The dominating set reconfiguration problem is defined as determining,
for a given dominating set problem and two among its feasible
solutions, whether one is reachable from the other via a sequence of
feasible solutions subject to a certain adjacency relation.
This problem is PSPACE-complete in general.
The concept of the dominating set is known to be quite useful for
analyzing wireless networks, social networks, and sensor networks.
We develop an approach \soh{3-1}{to solving}{to solve} the dominating set reconfiguration
problem based on Answer Set Programming (ASP).
Our declarative approach relies on a high-level ASP encoding, and both
the grounding and solving tasks are delegated to an ASP-based
combinatorial reconfiguration solver.
To evaluate the effectiveness of our approach, we conduct experiments
on a newly created benchmark set.

 \end{abstract}


\section{Introduction}\label{sec:introduction}

\emph{Combinatorial reconfiguration}~\cite[]{heuvel13,itdehapasiueun11,nishimura18}
is a relatively young research field of increasing interest in an
international community,
as witnessed by the international series of CoRe workshops.
The aim of combinatorial reconfiguration is to analyze
the structure and properties
of the solution spaces of combinatorial problems.
The typical topics of this field include the
reachability, optimality, diameter, and connectivity
of the solution spaces.
Each solution space has a graph structure,
in which each node represents an individual feasible solution, and
edges represent a certain adjacency relation. 
The reachability problem of combinatorial reconfiguration is defined
as the task of determining, 
for a given combinatorial problem and two of its feasible solutions,
whether there exists a sequence of adjacent feasible solutions from
one to another.
We refer to this problem as the \emph{Combinatorial Reconfiguration Problem} (CRP). 

\soh{1-1}{}{The study of combinatorial reconfiguration problems draws its motivation
from a variety of fields such as
statistical physics~\cite[]{mohsal09},
combinatorics~\cite[]{knuth11},
puzzles~\cite[]{Headem09},
industrial applications~\cite[]{itwkyktmh14}, and many others.
For illustration, the Potts model in physics is closely related to
graph coloring reconfiguration under an adjacency relation called Kempe change.
One motivation for combinatorial reconfiguration research is of
theoretical nature.
Combinatorial reconfiguration plays an important role in proving
the (parameterized) computational complexity of reconfiguration
counterparts of many central combinatorial problems.
Another motivation is very practical.
Reconfigurations are needed in many mission-critical applications.
For instance, in power distribution networks~\cite[]{itwkyktmh14},
we need to find a sequence of feasible switch configurations from one
to another without causing any blackout.
Regarding adjacency relations, in many cases on graph problems, 
the simplest relations (e.g., token jumping and token sliding) have
been studied in the literature.
But, in general, adjacency relations originate from applications.}

A solid theoretical foundation for combinatorial reconfiguration
problems has been established over the last decade.
Particularly, for many NP-complete problems, their reconfiguration
\soh{2-5}{problems}{counterparts} have been proven to be PSPACE-complete~\cite[]{itdehapasiueun11}.
Examples include
SAT reconfiguration \cite[]{gokomapa09,monipara17}, 
independent set reconfiguration \cite[]{itdehapasiueun11,kamemi12}, 
graph coloring reconfiguration \cite[]{boncer09,brmcmono16,cehejo11}, 
clique reconfiguration \cite[]{itonot15}, 
Hamiltonian cycle reconfiguration \cite[]{takaoka18}, 
set cover reconfiguration \cite[]{itdehapasiueun11},
and many others.
Very recently, starting with a series of international competitions
(CoRe challenge 2022 and 2023 \cite[]{sotaokit24}),\footnote{\url{https://core-challenge.github.io/2022/} and
  \url{https://core-challenge.github.io/2023/}} there \soh{2-5}{is}{has been} a growing interest in the practical aspects of
combinatorial reconfiguration problems~\cite[]{cherkamupeposesisp23,itkanasosuteto23,tiksut23}.
  
\emph{Dominating Set Reconfiguration Problems} (DSRP; \cite[]{Haasey14,bodoou21,haitmonionsute16,sumoni16})
is one of most theoretically studied combinatorial reconfiguration problems.
This problem is based on the well-known \emph{Dominating Set Problem} (DSP).
For a graph $G=(V,E)$, a subset of nodes $S \subseteq V$ is called
a \emph{dominating set} of $G$ when the union of $S$ and the set of
adjacent nodes to $S$ equals $V$. In other words, any $v\in
V\backslash S$ is adjacent to at least one node in $S$.
The task of DSP is to decide, for a given graph $G$ and a positive
integer $k$, whether there exists a dominating set of size $k$.
In general, \soh{2-5}{the dominating set problem}{DSP} is NP-complete,
and its reconfiguration (i.e., DSRP) is PSPACE-complete.
From a practical viewpoint, 
the dominating set problem has been well explored
since the concept of the dominating set is quite useful for analyzing
wireless networks, social networks, and sensor networks~\cite[]{bdtc05}.
However, little attention has been paid so far to the practical
aspects of dominating set reconfiguration.

In this paper, we present an approach \soh{3-1}{to solving}{to solve}
the dominating set reconfiguration problem based on
Answer Set Programming (ASP; \cite[]{lifschitz19a}).
In our approach, a DSRP instance is first converted into ASP facts.
\soh{2-5}{And then}{Then}, these facts are combined with a collection of ASP encodings
for DSRP solving, which are afterwards solved by the ASP-based CRP
solver {\recongo}.\footnote{\url{https://github.com/banbaralab/recongo}}
Clearly, our declarative approach based on ASP has several advantages.
The basic language of ASP is expressive enough for modeling a wide
range of combinatorial (optimization) problems.
\soh{3-2}{}{Recent advances in ASP indicate a promising direction to extend ASP to
be more applicable to dynamic problems. In particular, multi-shot ASP
solving allows for incremental grounding and solving for logic
programs in an operative way.
}
The language constructs of multi-shot ASP solving (e.g.,
\code{#program}) allow for an easy extension to their reconfiguration problems.
The {\recongo} solver, utilizing {\clingo}'s Python API for multi-shot ASP solving,
provides efficient reachability checking for
combinatorial reconfiguration problems.

The contributions and results of this paper are summarized as follows:
\begin{enumerate}
\item For the first step toward efficient DSRP solving,
  we compare two traditional ASP encodings for solving
  the minimum dominating set problem\soh{2-1}{.}{, an optimization
    variant of DSP that finds a dominating set of the minimum size.}
  We observed that one encoding
  used in ASP competition 2009~\cite[]{dvbgt09}
  performs well \soh{2-5}{compared with}{compared to} another \cite[]{Huynh20}.
  
\item We present an ASP encoding for DSRP solving under
  an adjacency relation called \emph{token jumping}.
  We also present a hint constraint on token destination
  to boost the performance of
  DSRP solving.
  Furthermore, we extend our encoding to dominating set
  reconfiguration under \emph{token addition-removal}.

\item We create, to our best knowledge, the first benchmark set of
  the dominating set reconfiguration problem under token jumping.
  The benchmark set consists of 442 instances in which
  310 are reachable and 132 are unreachable.
  The number of nodes and edges ranges from 11 to 1,000 and
  from 20 to 449,449, respectively.

\item Our ASP encoding manages to decide the reachability of 363 out
  of 442 instances.
  Particularly, our hint constraint can be highly effective in
  deciding unreachability.
  Furthermore, we establish the competitiveness of our declarative
  approach by empirically contrasting it to a more algorithmic
  ZDD-based approach~\cite[]{itkanasosuteto23}.
\end{enumerate}
Overall, our declarative approach can make a significant
contribution not only to the state-of-the-art of dominating set
reconfiguration, but also ASP application to combinatorial
reconfiguration.
This paper assumes some familiarity with ASP and its semantics as well
as multi-shot ASP solving~\cite[]{gekakasc17a}.
ASP encodings are written in the language of {\clingo}~\cite[]{PotasscoUserGuide19}.

 \section{Background}\label{sec:background}

The dominating set reconfiguration problem is defined as
the task of deciding,
for a given DSP instance and two among its feasible solutions $X_s$
and $X_g$,
whether there exists a sequence of transitions:
$X_{s}=X_{0}\rightarrow X_{1}\rightarrow X_{2}\rightarrow\cdots\rightarrow X_{\ell}=X_{g}$.
Each state $X_{i}$ represents a feasible solution (viz., dominating set).
We refer to $X_s$ and $X_g$ as the start and the goal states, respectively.
We write $X\rightarrow X'$ if state $X$ at step $t$ is followed
by state $X'$ at step $t+1$ under a certain \emph{adjacency relation}.
We refer to the sequence from $X_s$ to $X_g$ as a \emph{reconfiguration sequence}.
The \emph{length} of the reconfiguration sequence, denoted by $\ell$,
is the number of transitions.

In the literature, three kinds of adjacency relations have been well
studied in combinatorial reconfiguration~\cite[]{headem05,kamemi12,itdehapasiueun11}.
Suppose that a \emph{token} is placed on each node in a dominating set.
The \emph{token jumping} of $X\rightarrow X'$ means that 
a single token jumps from
the single node in $X \setminus X'$ to the one in $X' \setminus X$.
The \emph{token sliding} is a limited version of token jumping in which
a token slides along an edge.
The \emph{token addition-removal} means that
a single token is added or removed in each $X\rightarrow X'$
as long as the total number of tokens in $X_{i}$
does not exceed a given threshold.
In this paper, we mainly focus on token jumping.

\begin{figure}[t]
\centering
  \begin{tabular}{ccccc}
    \scalebox{0.80}{\begin{tikzpicture}
 \node[draw, circle] (A) at (0,2)    {1};
 \node[draw, fill=yellow,circle] (B) at (2,2)  {2};
 \node[draw, circle] (C) at (4,2)    {3};
 \node[draw, circle] (D) at (0,0)  {4};
 \node[draw, fill=yellow,circle] (E) at (2,0)  {5};
 \node[draw, circle] (F) at (4,0)  {6};
 \draw(A)--(B);
 \draw(A)--(D);
 \draw(B)--(C);
 \draw(B)--(E);
 \draw(C)--(E);
 \draw(C)--(F);
 \draw(D)--(E);
 \draw(E)--(F);
\end{tikzpicture} } & \smash{\lower-7.ex\hbox{$\rightarrow$}} &
    \scalebox{0.80}{\begin{tikzpicture}
 \node[draw, circle] (A) at (0,2)    {1};
 \node[draw, circle] (B) at (2,2)  {2};
 \node[draw, circle] (C) at (4,2)    {3};
 \node[draw, fill=yellow,circle] (D) at (0,0)  {4};
 \node[draw, fill=yellow,circle] (E) at (2,0)  {5};
 \node[draw, circle] (F) at (4,0)  {6};
 \draw(A)--(B);
 \draw(A)--(D);
 \draw(B)--(C);
 \draw(B)--(E);
 \draw(C)--(E);
 \draw(C)--(F);
 \draw(D)--(E);
 \draw(E)--(F);
\end{tikzpicture} } & \smash{\lower-7.ex\hbox{$\rightarrow$}} &
    \scalebox{0.80}{\begin{tikzpicture}
 \node[draw, circle] (A) at (0,2)    {1};
 \node[draw, circle] (B) at (2,2)  {2};
 \node[draw, fill=yellow,circle] (C) at (4,2)    {3};
 \node[draw, fill=yellow,circle] (D) at (0,0)  {4};
 \node[draw, circle] (E) at (2,0)  {5};
 \node[draw, circle] (F) at (4,0)  {6};
 \draw(A)--(B);
 \draw(A)--(D);
 \draw(B)--(C);
 \draw(B)--(E);
 \draw(C)--(E);
 \draw(C)--(F);
 \draw(D)--(E);
 \draw(E)--(F);
\end{tikzpicture} }\\
    $X_{0}$ (start state) & & {$X_{1}$} & & {$X_{2}$ (goal state)}\\    
  \end{tabular}
  \caption{An example of DSRP under token jumping}
  \label{fig:dsrp_tj}
\end{figure}
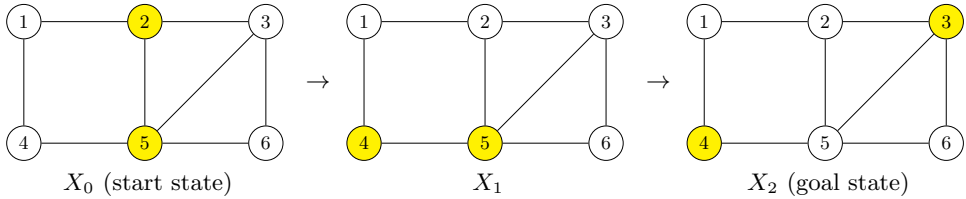

An example of the dominating set reconfiguration problem
under token jumping is shown in Figure \ref{fig:dsrp_tj}.
This example consists of a graph having 6 nodes and 8 edges,
and the size of dominating sets is $k=2$.
The dominating sets (i.e., tokens) are highlighted in yellow.
We can see that the goal state is reached from the start state
via a sequence of length $\ell = 2$.
Each state $X_{i}$ satisfies the constraints of dominating set problem.
In each transition, a single token moves under token jumping.
For example, 
in the transition from $X_{0}$ to $X_{1}$, a token jumps from node 2
in $X_{0}$ to node 4 in $X_{1}$.

 \section{Comparison of traditional ASP encodings for minimum DSP finding}
\label{sec:experiment}

The Dominating Set Reconfiguration Problem (DSRP) is based on the
Dominating Set Problem (DSP).
Therefore, for the first step toward efficient DSRP solving,
we compare two traditional ASP encodings for solving the minimum DSP. 
The task of this problem is to find, for a given graph, a dominating
set of the minimum size.

{\it The {\basicone} encoding} is shown in Listing~\ref{dsp1.lp}.
This encoding is a simplified version of one used in ASP competition
2009.\footnote{\url{https://dtai.cs.kuleuven.be/events/ASP-competition/encodings.shtml}}
Suppose that the nodes and edges are represented by the predicates 
\code{node/1} and \code{edge/2}, respectively.
The atom \code{in(X)} (cf., Line 1) is intended to represent that the
node \code{X} is in a dominating set, and characterizes a solution.
The auxiliary atom \code{dominated(X)}, introduced in Lines 2--3,
represents that the node \code{X} is either in a dominating set or
adjacent to at least one node in it.
The rule in Line 4 enforces that \code{dominated(X)} holds for every node \code{X}.
Finally, the number of nodes in a dominating set is minimized in
Line 5.
{\it The {\basictwo} encoding}~\cite[]{Huynh20} is shown in Listing~\ref{dsp3.lp}.
The difference from the {\basicone} encoding lies in the rule in Line 2.
That rule enforces that every node \code{X} not in a dominating set
is adjacent to at least one node in it.

\begin{lstlisting}[float=t,caption={The {\basicone} encoding: a simplified version of one used in ASP competition 2009},label=dsp1.lp,basicstyle=\ttfamily\small]
{ in(X) : node(X) }.
dominated(X) :- in(X).
dominated(Y) :- in(X), edge(X,Y).   dominated(Y) :- in(X), edge(Y,X).
:- not dominated(X), node(X).
#minimize { 1,X: in(X) }.
\end{lstlisting} 
\begin{lstlisting}[float=t,caption={The {\basictwo} encoding~{\cite[]{Huynh20}}},label=dsp3.lp,basicstyle=\ttfamily\small]
{ in(X) : node(X) }.
:- not 1 { in(Y):edge(X,Y) ; in(Y):edge(Y,X) }, not in(X), node(X).
#minimize { 1,X: in(X) }.
\end{lstlisting}
\begin{table}[t]
  \centering
  \tabcolsep = 5mm
  \renewcommand{\arraystretch}{1.0}
  \caption{Comparison results of the {\basicone} and {\basictwo} encodings}
  \begin{tabular}[t]{lrrrr}
  \topline
  & \multicolumn{2}{r}{{\basicone} encoding} &
  \multicolumn{2}{r}{{\basictwo} encoding} \\ 
  & \code{bb} & \code{usc} & \code{bb} & \code{usc} \midline
  \#optimal solutions & 46 & \textbf{124} & 47  & 123 \\ 
  \#unique solutions & \hspace{2.5em}32 & 12 & \hspace{2.5em}15 & 4 \midline
  total & 78 & \textbf{136} & 62 & 127 \botline
\end{tabular}

   \label{tab:opt.tex}
\end{table}

We carry out experiments to compare the performance of the {\basicone} and
{\basictwo} encodings.
Our empirical analysis considers all 167 graph instances used in CoRe
Challenge 2022, which are publicly available from the web.\footnote{\url{https://core-challenge.github.io/2022/}}
The number of nodes ranges from 11 to 40,000, and its average is 980.
We use the ASP solver {\clingo}-5.5.0 (default configuration) with 
two optimization strategies:
branch-and-bound (\code{bb}) and unsatisfiable core (\code{usc}).
The time-limit is 20 minutes for each instance.
We run our experiments on a Mac OS with a 3.2GHz Intel Core i7
processor and 64GB memory.

Comparison results are shown in Table~\ref{tab:opt.tex}.
The columns display the number of optimal and ``unique'' solutions for
each encoding.
A solution for an encoding is called unique if
there is \soh{2-5}{no other encoding, which}{no other encoding which} has a better objective value
or proves optimality for the same value.
The {\basicone} encoding with the \code{usc} optimization solved the most,
namely $136=124+12$ instances in which 124 are optimal and 12 are unique.
It is followed by $127=123+4$ of the {\basictwo} encoding with \code{usc}.
The optima obtained by the {\basicone} encoding with \code{usc} contain
the ones by the others.
Only one optimum obtained by the {\basicone} encoding with \code{usc}
but not by the others is for 
the \code{myciel7} instance, which has 191 nodes and 2,360 edges.
We can observe that the \code{usc} optimization performs well compared
to \code{bb} for both encodings.
In terms of CPU time of finding optima, 
the {\basicone} encoding with \code{usc} was able to find 124 optima
in 21.2 seconds on average.

Overall, as can be seen in Table~\ref{tab:opt.tex}, 
the {\basicone} encoding performed well \soh{2-5}{compared with}{compared to} the {\basictwo} encoding.
Although both encodings can be used, because of its efficiency, 
we will extend the {\basicone} encoding to dominating set
reconfiguration in next section.

 \section{ASP-based approach to dominating set reconfiguration}\label{sec:proposal}

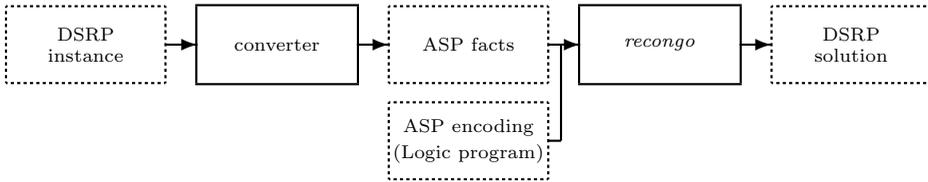
\begin{figure}[t]
    \centering
    \thicklines
    \setlength{\unitlength}{1.2pt}
    \small\footnotesize\scriptsize
    \begin{picture}(280,57)(4,-10)
      \put(  0, 20){\dashbox(50,24){\shortstack{DSRP\\ instance}}}
      \put( 60, 20){\framebox(50,24){converter}}
      \put(120, 20){\dashbox(50,24){\shortstack{ASP facts}}}
      \put(120,-10){\dashbox(50,24){\shortstack{ASP encoding\\(Logic program)}}}
      \put(180, 20){\framebox(50,24){\recongo}}
      \put(240, 20){\dashbox(50,24){\shortstack{DSRP\\solution}}}
      \put( 50, 32){\vector(1,0){10}}
      \put(110, 32){\vector(1,0){10}}
      \put(170, 32){\vector(1,0){10}}
      \put(230, 32){\vector(1,0){10}}
      \put(170, +2){\line(1,0){4}}
      \put(174, +2){\line(0,1){30}}
    \end{picture}  
  \caption{The architecture of our approach}
  \label{fig:arch}
  \end{figure}
\begin{lstlisting}[float=t,caption={ASP facts of a DSRP instance in Figure~\ref{fig:dsrp_tj}},numbers=none,label=code:dsrp_fact.lp,basicstyle=\ttfamily\small]
node(1).  node(2).  node(3).  node(4).  node(5).  node(6).
edge(1,2).  edge(1,4).  edge(2,3).  edge(2,5).
edge(3,5).  edge(3,6).  edge(4,5).  edge(5,6).
k(2).  start(2).  start(5).  goal(3).  goal(4).
\end{lstlisting}
\begin{lstlisting}[float=t,
  caption={ASP encoding for DSRP solving under token jumping},label=code:jump.lp,basicstyle=\ttfamily\small]
#program base.
% Constraints of the start state
:- not in(X,0), start(X).

#program step(t).
% Constraints of the dominating set problem
K { in(X,t) : node(X) } K :- k(K).
dominated(X,t) :- in(X,t).
dominated(Y,t) :- in(X,t), edge(X,Y).
dominated(Y,t) :- in(X,t), edge(Y,X).
:- not dominated(X,t), node(X).

% Constraints of the token jumping
token_removed(X,t) :- in(X,t-1), not in(X,t), t > 0.
:- not 1 { token_removed(X,t) } 1, t > 0.

#program check(t).
% Constraints of the goal state
:- not in(X,t), goal(X), query(t).
\end{lstlisting} 
\begin{lstlisting}[float=t,
  caption={Token destination: a hint constraint for DSRP solving},label=code:hint_token3_inc.lp,numbers=none,basicstyle=\ttfamily\small]
#program step(t).
:- not 1 { in(Y,t):edge(X,Y) ; in(Y,t):edge(Y,X) } 1,
	 { in(Y,t-1):edge(X,Y) ; in(Y,t-1):edge(Y,X) } 0,
	 token_removed(X,t), t>0.
\end{lstlisting} 
\begin{figure}[t]\centering
  \tabcolsep = 4mm  
 \begin{tabular}{ccc}
 {$X_{t-1}$} & & {$X_{t}$} \\
 \scalebox{0.80}{\begin{tikzpicture}
 \tikzset{wc/.style={circle,draw=white,minimum size=0.4cm}}
 \node[draw, fill=yellow,circle] (A) at (1,0.8)    {1};
 \node[draw, circle] (B) at (1,2)  {2};
 \node[draw, circle] (C) at (0,0)    {3};
 \node[draw, circle] (D) at (2,0)  {4};
 \node[draw, circle] (E) at (3,2)  {5};
 \draw(A)--(B);
 \draw(A)--(C);
 \draw(A)--(D);
\end{tikzpicture} } & 
 \smash{\lower-7.ex\hbox{$\rightarrow$}} & 
 \scalebox{0.80}{\begin{tikzpicture}
 \tikzset{wc/.style={circle,draw=white,minimum size=0.4cm}}
 \node[draw, circle] (A) at (1,0.8)    {1};
 \node[draw, fill=yellow,circle] (B) at (1,2)  {2};
 \node[draw, circle] (C) at (0,0)    {3};
 \node[draw, circle] (D) at (2,0)  {4};
 \node[draw, fill=yellow,circle] (E) at (3,2)  {5};
 \draw(A)--(B);
 \draw(A)--(C);
 \draw(A)--(D);
 \draw[thick, dashed, -latex] (A) to[bend right=25] (E);
 \draw[thick, dashed, -latex] (A) to[bend left=60] (B);
 \node at (2.2,1.4) {\large\color{red}\KO};
 \node at (0.2,1.35) {\large\color{red}\OK};
\end{tikzpicture} } 
 \end{tabular}
 \caption{Example of invalid move forbidden by the hint on token destination},\label{fig:hint_t3}
\end{figure}
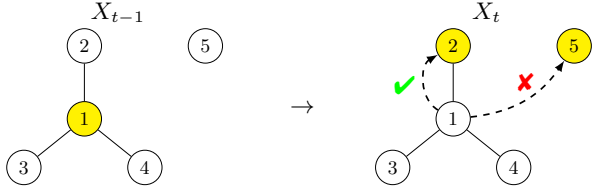

We now present an approach \soh{3-1}{to solving}{to solve} the dominating set
reconfiguration problem based on ASP.
The architecture of our approach is shown in Figure~\ref{fig:arch}.
The resulting solver accepts a DSRP instance in DIMACS format and
converts it into ASP facts.
\soh{2-5}{And then these facts}{And then, these facts} are combined with a collection of ASP encodings
for DSRP solving, which are afterwards solved by the {\recongo} solver.
{\recongo} is an ASP-based CRP solver powered by {clingo}'s multi-shot ASP solving.

The input of DSRP consists of a DSP instance and two among its
feasible solutions (the start and goal states).
ASP facts of the example in Figure~\ref{fig:dsrp_tj} are shown in 
Listing~\ref{code:dsrp_fact.lp}.
The predicates \code{node/1} and \code{edge/2} represent the nodes and
edges, respectively.
The size of dominating sets is represented by the predicate \code{k/1}.
The predicates \code{start/1} and \code{goal/1}
represent the start and goal states, respectively.
In this example, the dominating set of the start state is \{2,5\}
since \code{start(2)} and \code{start(5)} are given.

\textbf{First-order encoding. }
Our ASP encoding for DSRP solving under token jumping
is shown in Listing~\ref{code:jump.lp}.
This encoding consists of three subprograms
indicated by \code{#program} statements:
\code{base}, \code{step(t)}, and \code{check(t)}.
The parameter \code{t} is a symbolic constant that represents a 
step in reconfiguration sequences.
\code{base} is a default subprogram and includes all rules that are
not preceded by any other \code{#program} statements.
The atom \code{in(X,t)} (cf., Line 7) represents that
the node \code{X} is in a dominating set at step \code{t}, and
characterizes a solution of DSRP.
The \code{base} subprogram specifies constraints
to be satisfied in the start state.
The rule in Line 3 enforces that the node \code{X} is in a dominating
set at step \code{0} if \code{start(X)} holds.
The \code{step(t)} subprogram specifies constraints to be satisfied at
each step \code{t}.
The rules in Lines 7--11 represent the constraints of the dominating
set problem. 
The rule in Line 14 introduces the auxiliary atom
\code{token_removed(X,t)}, which represents
that a token is removed from node \code{X} at step \code{t}.
The rule in Line 15 enforces that 
single token jumps (i.e., is removed) from the single node in each
step. The \code{check(t)} subprogram represents constraints to be satisfied
in the goal state.
The rule in Line 19 checks whether or not 
the goal state is reached at step \code{t}.
The activation or deactivation of this rule is
controlled by the external atom \code{query(t)},
whose truth value can be changed later.

\textbf{Reachability checking with {\recongo}. }
For a given DSRP instance $I$ in fact format, 
{\recongo} constructs a logic program 
\(
  \varphi_{\ell} =
  I
  \cup
  \texttt{base}
  \cup
  \bigcup_{\texttt{t}=0}^{\ell} \texttt{step(t)}
  \cup
  \texttt{check(}\ell\texttt{)}
\).
Here, \texttt{base}, \texttt{step(t)}, \texttt{check($\ell$)}
correspond to the three subprograms
given in Listing~\ref{code:jump.lp}, respectively.
We note that $\varphi_{\ell}$ represents the dominating set
reconfiguration problem of a bounded length $\ell$.
For solving $\varphi_{\ell}$, {\recongo}
delegates the grounding and solving tasks to the {\clingo} solver.
If $\varphi_{\ell}$ is satisfiable, there exists a reconfiguration sequence.
Otherwise, {\recongo} keeps on reconstructing a successor
(e.g., $\varphi_{\ell+1}$) and checking its satisfiability until a
reconfiguration sequence is found.
Of course, it is quite inefficient to fully reconstruct
$\varphi_{\ell}$ at each step due to the expensive grounding.
To resolve this issue, 
{\recongo} incrementally constructs $\varphi_{\ell}$ from its 
predecessor $\varphi_{\ell-1}$ by adding all rules of 
\texttt{step($\ell$)} and \texttt{check($\ell$)}, and
by deactivating \texttt{check($\ell$-1)} by
setting the external atom \texttt{query($\ell-1$)} to false.
This procedure can be elegantly implemented by utilizing {\clingo}'s
API for multi-shot ASP solving~\cite[]{gekakasc17a}.

\textbf{Hint constraints. }
We develop a hint constraint,
called \emph{token destination},
to boost the performance of DSRP solving.
This domain-specific hint is intended to forbid invalid token moves.
Our ASP encoding for the token destination is shown in Listing~\ref{code:hint_token3_inc.lp}.
The rule enforces that a token moves from the node \code{X} to one among its
neighbors if the token is removed from the node \code{X} at step \code{t}
and no token is placed on all its neighbors at step \code{t-1}.
For illustration, an invalid move forbidden by this hint is shown in 
Figure~\ref{fig:hint_t3}.
Indeed, node 1 is neither in a dominating set nor adjacent to any node
in it if a token jumps from node 1 in $X_{t-1}$ to 5 in $X_{t}$.

\textbf{Extension. }
Finally, we extend our encoding to DSRP solving under token addition-removal.
In the token addition-removal, 
a single token is added or removed in each transition
as long as the total number of tokens in each state does not exceed a
given threshold.
An extended encoding is shown in Listing~\ref{code:tar.lp}.
The major difference from token jumping in
Listing~\ref{code:jump.lp} is that
the auxiliary atom \code{token_added(X,t)} is newly introduced
in Line 16.
The atom \code{token_added(X,t)} represents that a token is added to
node \code{X} at step \code{t}.
The rule in Line 17 enforces that 
a single token is either added or removed in each transition.
For minor changes, we add a threshold \code{th} in Line 8
to bind the number of tokens (viz., the size of dominating sets).
Since there is no lower bound,
the constraints of the start and goal states are adjusted.

\begin{lstlisting}[float=t,
  caption={ASP encoding for DSRP solving under token addition-removal},label=code:tar.lp,basicstyle=\ttfamily\small]
#program base.
% Constraints of the start state
:- not in(X,0), start(X).
:- in(X,0), not start(X).

#program step(t).
% Constraints of the dominating set problem
{ in(X,t) : node(X) } K + th :- k(K).
dominated(X,t) :- in(X,t).
dominated(Y,t) :- in(X,t), edge(X,Y).
dominated(Y,t) :- in(X,t), edge(Y,X).
:- not dominated(X,t), node(X).

% Constraints of the token addition-removal
token_removed(X,t) :- in(X,t-1), not in(X,t), t > 0.
token_added(X,t) :- not in(X,t-1), in(X,t), t > 0.
:- not 1 { token_removed(X,t) ; token_added(X,t) } 1, t > 0.

#program check(t).
% Constraints regarding of goal state
:- not in(X,t), goal(X), query(t).
:- in(X,t), not goal(X), query(t).
\end{lstlisting}

 \section{Benchmark generation and Experiments}\label{sec:experiment_dsrp}

We carry out experiments on a newly created benchmark set to evaluate
the effectiveness of our approach.
We compare the combination of our ASP encoding
and hint constraints for DSRP solving under token jumping.
We also address the competitiveness of our approach by
empirically contrasting it to a ZDD-based approach~\cite[]{itkanasosuteto23}.

\textbf{Benchmark generation. }
No benchmark set of DSRP has been available so far.
We therefore create a benchmark set of DSRP under token jumping.
The resulting benchmark set consists of 442 instances in total,
in which 310 are reachable, and 132 are unreachable.
The number of nodes and edges ranges from 11 to 1,000 and
from 20 to 449,449, respectively.
All benchmark instances are in DIMACS format used in 
a series of international competitions
(CoRe challenge), and can be useful for other approaches
and solvers.
More precisely, our benchmark set is generated as follows:
\begin{enumerate}
\item \soh{2-2}{We ran experiments for enumerating the optimal solutions
  of the minimum DSP.
  We considered all 167 instances (viz., graphs)
  used in the third benchmark set of CoRe Challenge 2022}{We considered all 167 instances (viz., graphs)
  used in the third benchmark set of CoRe Challenge 2022. We ran experiments for enumerating the optimal solutions
  of the minimum DSP}.~\footnote{This benchmark set is the largest one involving many kinds of graphs.
    The first set consists of toy instances.
    The second set focuses on the grid and queen graphs.
    See \url{https://core-challenge.github.io/2022/} for more details.}
  The time-limit is 5 minutes for each instance.
  {\clingo} was able to fully enumerate the optimal solutions of 57 instances.
\item For each of the 57 instances,
  we tried to construct its solution space using breadth-first search.
  A solution space is a graph in which each node represents a
  dominating set, and each edge represents an adjacency relation,
  in our case, the token jumping.
  We obtained the solution spaces of 40 instances.
\item For each solution space of the 40 instances, we tried to
  produce both at most 10 reachable instances and at most 10
  unreachable ones. 
  Finally, we were able to create 442 instances in a total, of which
  310 are reachable and 132 are unreachable.
  \label{benchmark:step:3}
\end{enumerate}
In step (\ref{benchmark:step:3}), for reachable instances, 
the start and goal states are selected for which the length of the
shortest sequence becomes maximum (viz., diameter).
The lengths of reachable instances range from 1 to 23.
For unreachable instances,
the start and goal states are selected from different connected
components of the solution space respectively.

\textbf{Overview of experiments. }
We compare the combination of our ASP encoding
in Listing~\ref{code:jump.lp}
and hint constraints for DSRP solving under token jumping.
In addition to our hint constraint on token destination
(\code{t3}) in Listing~\ref{code:hint_token3_inc.lp},
we consider the following hint constraints.
\begin{itemize}
\item \emph{Distance from the start and goal states} (\code{d1} and \code{d2}):\\
The \code{d1} hint represents that there must be at most $t$ nodes
  that are in the start state but not in $X_{t}$.
  Let us consider a reconfiguration sequence of length $\ell$.
  The \code{d2} hint represents that there must be at most $\ell-t$
  nodes that are in the goal state but not in step $X_{t}$.
\item \emph{Forbidding redundant token moves} (\code{t1} and \code{t2}):\\
The \code{t1} hint represents that, in two consecutive states,
  no token moves back to a node from which a token moved before.
  Similarly, the \code{t2} hint represents that
  no token moves from a node to which a token moved before.
\item \emph{Minimal dominating set heuristics} (\code{heu}):\\
  The \code{heu} hint is a domain-specific heuristics that tries to
  make each state to be a minimal dominating set.
  This can be easily done by using {\clingo}'s \code{#heuristic}
  statements.
\end{itemize}
These hints except for \code{heu} can be used for a wide range of CRPs
and have been applied to
independent set reconfiguration~\cite[]{ybisu24} and 
Hamiltonian cycle reconfiguration~\cite[]{hibainlunascsota23}.
We use the ASP-based CRP solver {\recongo}-0.3 powered by
{\clingo}-5.6.2 (configuration \textit{trendy}).
The time-limit is 10 minutes for each instance.
We run our experiments on Mac OS with a 3.2GHz Intel Core i7 processor and 64GB of memory.

\begin{table}[t]\centering
  \caption{The number of solved instances for DSRP solving with single hint}
  \tabcolsep = 3mm
\begin{tabular}[t]{lrrrrrrr}
  \topline
  & \code{nohint} & \code{d1} & \code{d2} & \code{heu} & \code{t1} & \code{t2} & \code{t3} \midline
  reachable & 293 & 291 & 293 & 288 & 297 & {\bf 299} & 296  \\ 
  unreachable & 38 & 38 & 38 & 32 & 38 & 48 & {\bf 58}  \midline
  total & 331 & 329 & 331 & 320 & 335 & 347 & {\bf 354} \botline
\end{tabular}

   \label{tb:dsrp:existent:single}
\end{table}
\begin{table}[t]\centering
  \caption{The number of solved instances for DSRP solving with multiple hints}
  \tabcolsep = 3mm
\begin{tabular}[t]{lrrrrrrrr}\topline
  & \code{nohint} & \code{t1}  & \code{t2} & \code{t3} & \code{t1t2} & \code{t1t3} & \code{t2t3} & \code{t1t2t3} \midline
  reachable & 293    & 297 & {\bf 299}& 296 & 298& 297& 291 & 295\\ 
  unreachable & 38     & 38  & 48& 58& 48& 58& {\bf 69} & 68 \midline
  total & 331 & 335& 347& 354& 346& 355& 360 & {\bf 363}\botline
\end{tabular}

  \label{tb:dsrp:existent:multi}
\end{table}

\textbf{Benchmark results. }
We first analyze the effectiveness of single hint constraint.
The number of solved instances is shown in Table~\ref{tb:dsrp:existent:single}.
The columns show in order
reachability (reachable, unreachable, or total)
and the number of solved instances by each single hint.
\code{nohint} indicates our ASP encoding without any hint.
The best results are highlighted in bold.
Our \code{t3} hint solved the most, namely 354 out of 442 instances,
which is 23 more than \code{nohint}.
For reachable instances, the \code{t2} hint solved the most,
but no big difference from the others.
For unreachable instances,
our \code{t3} hint solved the most, namely 58 instances, which is 20 more than \code{nohint}.

Next, we consider all possible combinations of
\{\code{t1},\code{t2},\code{t3}\},
each of which solved more than \code{nohint}
in Table~\ref{tb:dsrp:existent:single}.
The number of solved instances for each combination is shown in
Table~\ref{tb:dsrp:existent:multi}.
The \code{t1t2t3} hints solved the most, namely 363 instances.
It is 32 more than \code{nohint} and 9 more than the best single hint \code{t3}.
For reachable instances, the \code{t2} hint solved the most as before,
but again no big difference from the others.
For unreachable instances, 
the \code{t2t3} hints solved the most, namely 69 instances.
It is 31 more than \code{nohint} and 11 more than the best single hint \code{t3}.

\begin{figure}[t]
\begin{minipage}{0.5\textwidth}\centering
  \includegraphics[width=\linewidth]{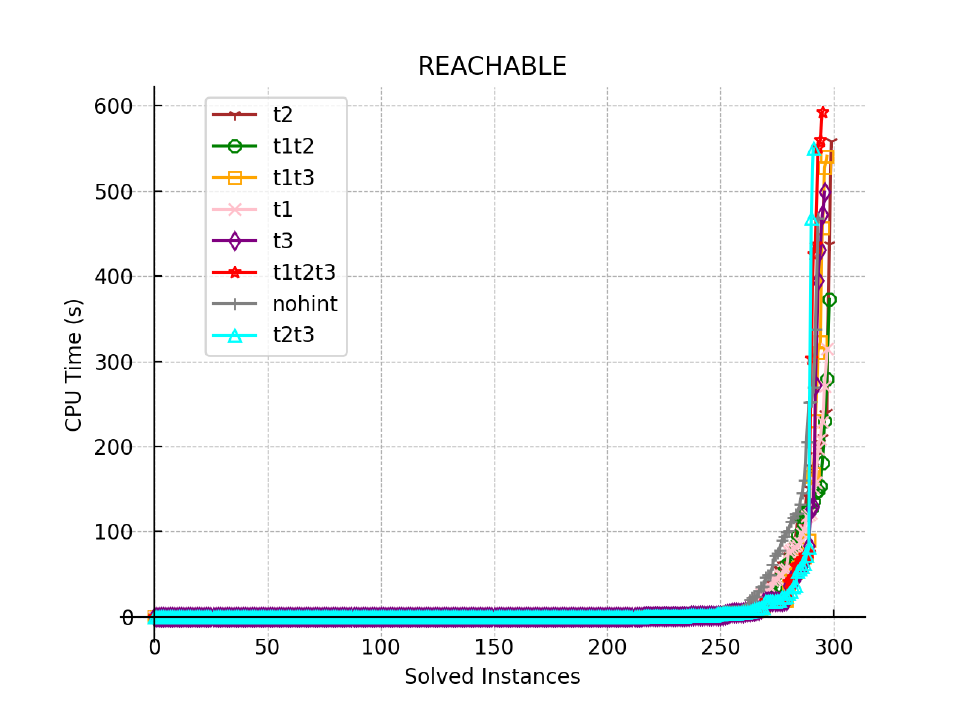}
  \caption{Cactus plot of reachable instances}
  \label{fig:cuctus_reach.png}
\end{minipage}\begin{minipage}{0.5\textwidth}\centering
  \includegraphics[width=\linewidth]{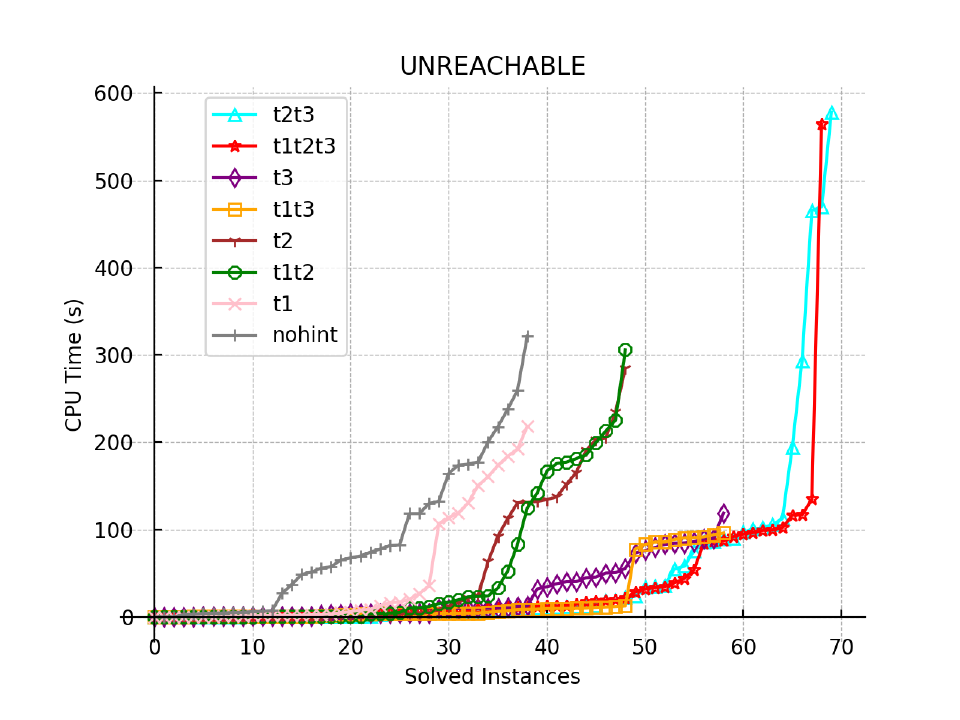}
  \caption{Cactus plot of unreachable instances}
  \label{fig:cuctus_unreach.png}
\end{minipage}
\end{figure}
\begin{figure}[t]\centering
  \includegraphics[width=0.6\linewidth]{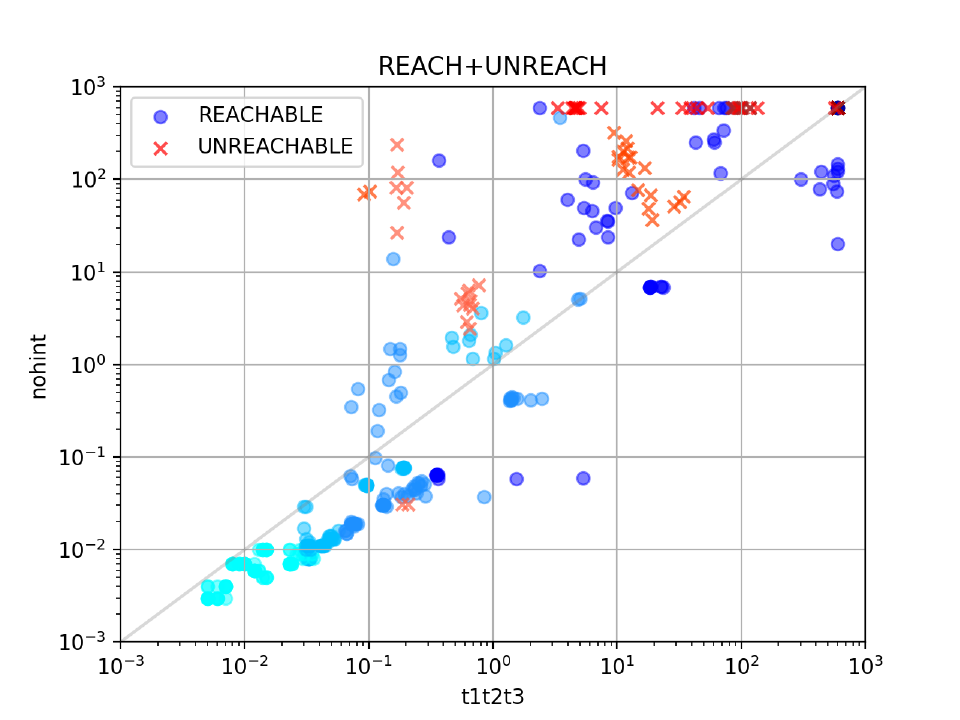}
  \caption{Scatter plot of CPU times comparing \code{nohint} and \code{t1t2t3}}
  \label{fig:scatter.png}
\end{figure}

We show cactus plots in 
Figures \ref{fig:cuctus_reach.png} and \ref{fig:cuctus_unreach.png} 
to visually summarize our benchmark results.
The graph plots how many instances can be solved in a given time-limit.
The horizontal axis indicates the number of solved instances, and 
the vertical axis indicates CPU times in seconds for reachability checking.
For unreachable instances, in Figure~\ref{fig:cuctus_unreach.png},
we can observe that there is a clear gap between
the top four encodings with \code{t3} and
the bottom four without it.
That is, this cactus plot illustrates that our \code{t3} hint can
significantly enhance the performance of deciding unreachability.

For a more detailed analysis of this effect,
a scatter plot of the comparison on CPU times between 
\code{nohint} and \code{t1t2t3} is shown in Figure~\ref{fig:scatter.png}.
Reachable and unreachable instances are indicated in blue and red, respectively.
The darker the color of each mark, the larger the ``scale'' of instance.
Here, the scale of instance is calculated by the product of the number
of nodes in the input graph and the length of the reconfiguration sequence.
The scatter plot shows that 
the \code{t1t2t3} hints are able to efficiently determine the
unreachability of many large instances colored by dark red
in the upper right part.

\begin{table}[t]
  \centering
  \tabcolsep = 5mm
  \renewcommand{\arraystretch}{1.0}
  \caption{Comparison with a ZDD-based approach}
  \begin{tabular}[t]{lrrr}
  \topline
  benchmark family & \#instances & ZDD ({\ddreconf})  & ASP (\code{t1t2t3}) \midline
  \code{color04} & 304 & 141 & {\bf 245} \\
  \code{queen}   &  66 &  {\bf 60} &  46 \\
  \code{sp}      &  60 &  {\bf 60} &  {\bf 60} \\
  \code{square}  &  12 &  {\bf 12} &  {\bf 12} \midline
  total   & 442 & 273 & {\bf 363} \botline
\end{tabular}

   \label{tb:ddreconf_vs_recongo}
\end{table}

\begin{figure}[t]
  \includegraphics[width=0.55\linewidth]{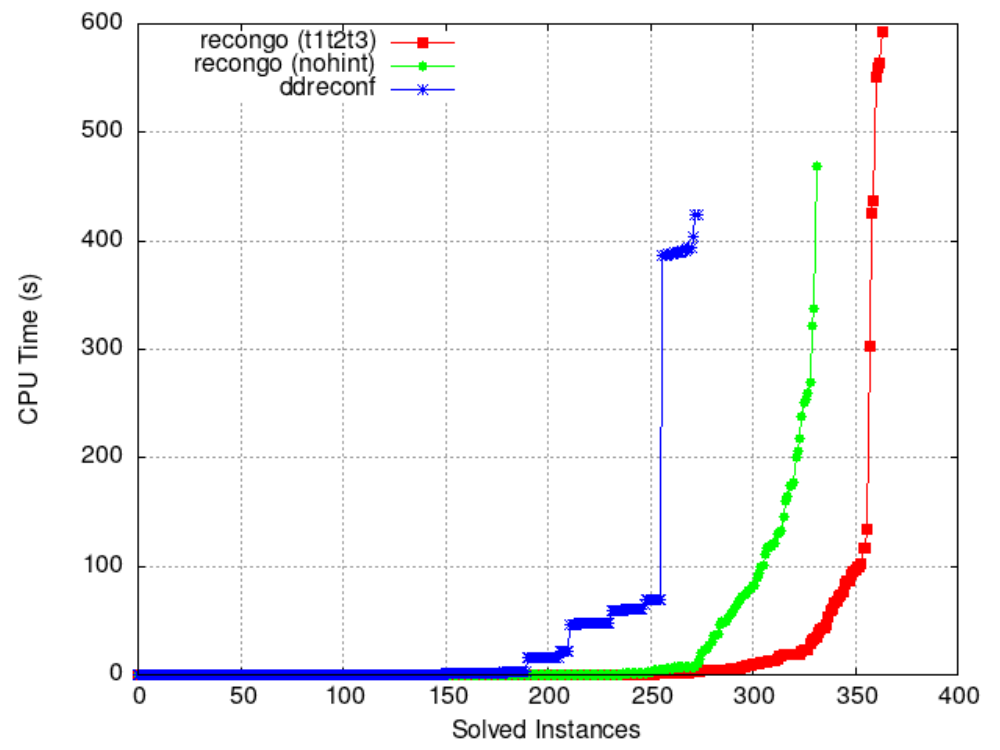}
  \caption{Cactus plot comparing ASP-based and ZDD-based approaches}
  \label{fig:cactus_recongo_ddreconf.png}
\end{figure}

\textbf{Comparison with other approaches.}
Our comparison considers a state-of-the-art approach to CRP solving
based on Zero-suppressed binary Decision Diagrams (ZDD; \cite[]{itkanasosuteto23}).
We use the ZDD-based CRP solver {\ddreconf}~\footnote{\url{https://github.com/junkawahara/ddreconf}}
ranked third in several metrics of the CoRe Challenge 2023.
For a given DSRP instance (i.e., a DSP instance and start and goal states),
{\ddreconf} first constructs a ZDD representing all solutions of DSP.
And then, it incrementally constructs ZDDs $Z^i$ representing 
all dominating sets that are reachable from the start state in exactly
$i$ steps, excluding ones that are reachable in less than $i$ steps. 
If $Z^i$ includes the goal state, \textit{ddreconf} returns a reconfiguration sequence.
We ran {\ddreconf} on the same environment as before.

Comparison results are shown in Table~\ref{tb:ddreconf_vs_recongo}.
The columns display in order
the benchmark family, the number of instances belonging to each family,
the number of solved instances for each approach.
The better results of the last two columns are highlighted in bold.
Overall, our declarative approach performed well \soh{2-5}{compared with}{compared to} a
ZDD-based approach.
Our ASP encoding with {\recongo} solved 363 out of 442 instances,
which is 90 more than {\ddreconf}.
Our encoding with {\recongo} showed good performance for the \code{color04}
family including many kinds of graphs, but less effective for
the \code{queen} family consisting of only graphs
for queens puzzles.
The cactus plot in Figure~\ref{fig:cactus_recongo_ddreconf.png}
illustrates a clear difference in performance between
ASP-based and ZDD-based approaches.

\soh{1-3}{}{\textbf{Summary and discussion.}
Our basic ASP encoding in Listing~\ref{code:jump.lp}
solved 331 instances without any hint constraints.
In contrast, our best encoding involving \code{t1t2t3} hints solved 
363 instances.
The improvement by ASP is therefore 331 instances, and 
the additional 32 instances are contributed by our hint constraints.
From the perspective of ZDD versus ASP in Figure~\ref{fig:cactus_recongo_ddreconf.png},
{\ddreconf} and {\recongo}(\code{nohint}) solved 273 and 331 instances, respectively.
This result shows that our ASP-based method is more effective in DSRP
solving than ZDD, even without the help of hint constraints.
}.

\soh{2-3}{}{The most recent competition (CoRe challenge 2023) has six metrics in
the single solver track.
The independent set reconfiguration problems under the token jumping
have been used in CoRe challenge 2023.
The {\recongo} solver won four gold and two silver medals, which is
followed by two gold and four silver medals of {\paris}~\cite[]{cherkamupeposesisp23}, 
and by six bronze medals of {\ddreconf}.
The gap between {\recongo} and {\ddreconf} at the competition is very
similar to the results on DSRPs in Table~\ref{tb:ddreconf_vs_recongo}.
}
We used the {\ddreconf} solver in our comparison, since,
among solvers that participated in a series of CoRe challenge,
only {\ddreconf} can deal with dominating set reconfiguration.
In principle, it is possible for other CRP solvers like {\paris}
to handle it, but significant efforts of high-level modeling and/or
encoding are needed.

 \section{Related work}\label{sec:related}

Recent advances in ASP such as
multi-shot ASP solving~\cite[]{gekakasc17a}
encourage researchers to tackle hard problems in combinatorial
reconfiguration.
\soh{2-4}{}{The use of multi-shot ASP for combinatorial reconfiguration was first
studied in \cite[]{yabainsc23,ybisu24}.
They proposed a general approach called bounded combinatorial
reconfiguration for solving combinatorial reconfiguration problems
based on ASP, including
algorithms, solver developments, encodings, and empirical analysis.
The resulting ASP-based solver {\recongo} is an award-winning solver
of the most recent CoRe challenge 2022 and 2023.
}

\soh{2-4}{}{The bounded combinatorial reconfiguration has been applied to some
specific reconfiguration problems.
\cite{yabainsc23,ybisu24} developed a collection of ASP encodings
for independent set reconfiguration under the token jumping rule,
including basic encodings for problem solving and some hint constraints.
\cite{hibainlunascsota23} explored Hamiltonian cycle reconfiguration
under the well-known k-opt rule of the traveling salesperson problem.
They presented new ASP encodings for solving Hamiltonian cycle
problems and extended one of them for Hamiltonian cycle
reconfiguration.
}

\soh{2-4}{}{This paper tackled the dominating set reconfiguration problem (DSRP).
The major contribution of this paper is the development of ASP
encodings and a hint constraint (named token destination) for
DSRP solving under token jumping as well as token addition-removal.
Particularly, our best hint constraint is shown to be highly effective
in deciding unreachability. This hint is domain-specific to DSRP and
is different from the ones used in the previous
works~\cite[]{yabainsc23,ybisu24,hibainlunascsota23}.
Our approach has a similarity to the previous works in the sense that
both the grounding and solving tasks are delegated to the {\recongo} solver.
}

There is a rapidly growing interest in the practical aspects of
combinatorial reconfiguration.
However, many important problems remain untouched, such as
the connectivity, optimality, and diameter of the solution space.
We will investigate the possibility of using ASP for those challenging
problems.

 \section{Conclusion}\label{sec:conclusion}

We developed an approach \soh{3-1}{to solving}{to solve} the dominating set reconfiguration
problem based on Answer Set Programming (ASP).
Our declarative approach relies on a high-level ASP encoding, and both
the grounding and solving tasks are delegated to the ASP-based
combinatorial reconfiguration solver {\recongo}.
We established the competitiveness of our declarative approach by
empirically contrasting it to a more algorithmic ZDD-based approach.
All source code and benchmark problems are available
from the web.~\footnote{\url{https://github.com/banbaralab/iclp2024}}

Future work includes benchmark generation for DSRP under
token addition-removal, using real-data of social and sensor networks. 
From a broader perspective, combinatorial reconfiguration is related
to automated planning, in the sense of transforming a given state to
another state. It would be interesting to study the relationship
between them and investigate the possibility of their synergy.
For a synergy, we plan to develop a framework of cost-optimal
combinatorial reconfiguration with multiple adjacency relations.

\section*{acknowledgments}

This work was partially supported by JSPS KAKENHI Grant Numbers
JP20H05964, JP22K11973, JP24H00686, and
DFG grant SCHA 550/15.

\bibliographystyle{acmtrans}

\begin{thebibliography}{}

\bibitem[\protect\citeauthoryear{Blum, Ding, Thaeler, and Cheng}{Blum
  et~al\mbox{.}}{2005}]{bdtc05}
{\sc Blum, J.}, {\sc Ding, M.}, {\sc Thaeler, A.}, {\sc and} {\sc Cheng, X.}
  2005.
\newblock Connected dominating set in sensor networks and manets.
\newblock In {\em Handbook of Combinatorial Optimization}, {D.-Z. Du} {and}
  {P.~M. Pardalos}, Eds. Springer, 329--369.

\bibitem[\protect\citeauthoryear{Bonamy, Dorbec, and Ouvrard}{Bonamy
  et~al\mbox{.}}{2021}]{bodoou21}
{\sc Bonamy, M.}, {\sc Dorbec, P.}, {\sc and} {\sc Ouvrard, P.} 2021.
\newblock Dominating sets reconfiguration under token sliding.
\newblock {\em Discrete Applied Mathematics\/}~{\em 301}, 6--18.

\bibitem[\protect\citeauthoryear{Bonsma and Cereceda}{Bonsma and
  Cereceda}{2009}]{boncer09}
{\sc Bonsma, P.~S.} {\sc and} {\sc Cereceda, L.} 2009.
\newblock Finding paths between graph colourings: {PSPACE}-completeness and
  superpolynomial distances.
\newblock {\em Theoretical Computer Science\/}~{\em 410,\/}~50, 5215--5226.

\bibitem[\protect\citeauthoryear{Brewster, McGuinness, Moore, and
  Noel}{Brewster et~al\mbox{.}}{2016}]{brmcmono16}
{\sc Brewster, R.~C.}, {\sc McGuinness, S.}, {\sc Moore, B.~R.}, {\sc and} {\sc
  Noel, J.~A.} 2016.
\newblock A dichotomy theorem for circular colouring reconfiguration.
\newblock {\em Theoretical Computer Science\/}~{\em 639}, 1--13.

\bibitem[\protect\citeauthoryear{Cereceda, van~den Heuvel, and
  Johnson}{Cereceda et~al\mbox{.}}{2011}]{cehejo11}
{\sc Cereceda, L.}, {\sc van~den Heuvel, J.}, {\sc and} {\sc Johnson, M.} 2011.
\newblock Finding paths between 3-colorings.
\newblock {\em Journal of Graph Theory\/}~{\em 67,\/}~1, 69--82.

\bibitem[\protect\citeauthoryear{Christen, Eriksson, Katz, Muise, Petrov,
  Pommerening, Seipp, Sievers, and Speck}{Christen
  et~al\mbox{.}}{2023}]{cherkamupeposesisp23}
{\sc Christen, R.}, {\sc Eriksson, S.}, {\sc Katz, M.}, {\sc Muise, C.}, {\sc
  Petrov, A.}, {\sc Pommerening, F.}, {\sc Seipp, J.}, {\sc Sievers, S.}, {\sc
  and} {\sc Speck, D.} 2023.
\newblock {PARIS:} planning algorithms for reconfiguring independent sets.
\newblock In {\em Proceedings of the 26th European Conference on Artificial
  Intelligence (ECAI 2023)}, {K.~Gal}, {A.~Now{\'{e}}}, {G.~J. Nalepa},
  {R.~Fairstein}, {and} {R.~Radulescu}, Eds. Frontiers in Artificial
  Intelligence and Applications, vol. 372. {IOS} Press, 453--460.

\bibitem[\protect\citeauthoryear{Denecker, Vennekens, Bond, Gebser, and
  Truszczynski}{Denecker et~al\mbox{.}}{2009}]{dvbgt09}
{\sc Denecker, M.}, {\sc Vennekens, J.}, {\sc Bond, S.}, {\sc Gebser, M.}, {\sc
  and} {\sc Truszczynski, M.} 2009.
\newblock The second answer set programming competition.
\newblock In {\em Proceedings of the 10th International Conference on Logic
  Programming and Nonmonotonic Reasoning ({LPNMR} 2009)}. LNCS, vol. 5753.
  Springer, 637--654.

\bibitem[\protect\citeauthoryear{Gebser, Kaminski, Kaufmann, Lindauer,
  Ostrowski, Romero, Schaub, Thiele, and Wanko}{Gebser
  et~al\mbox{.}}{2019}]{PotasscoUserGuide19}
{\sc Gebser, M.}, {\sc Kaminski, R.}, {\sc Kaufmann, B.}, {\sc Lindauer, M.},
  {\sc Ostrowski, M.}, {\sc Romero, J.}, {\sc Schaub, T.}, {\sc Thiele, S.},
  {\sc and} {\sc Wanko, P.} 2019.
\newblock {\em Potassco User Guide\/}, Version 2.2.0 ed.

\bibitem[\protect\citeauthoryear{Gebser, Kaminski, Kaufmann, and Schaub}{Gebser
  et~al\mbox{.}}{2019}]{gekakasc17a}
{\sc Gebser, M.}, {\sc Kaminski, R.}, {\sc Kaufmann, B.}, {\sc and} {\sc
  Schaub, T.} 2019.
\newblock Multi-shot {ASP} solving with clingo.
\newblock {\em Theory and Practice of Logic Programming\/}~{\em 19,\/}~1,
  27--82.

\bibitem[\protect\citeauthoryear{Gopalan, Kolaitis, Maneva, and
  Papadimitriou}{Gopalan et~al\mbox{.}}{2009}]{gokomapa09}
{\sc Gopalan, P.}, {\sc Kolaitis, P.~G.}, {\sc Maneva, E.~N.}, {\sc and} {\sc
  Papadimitriou, C.~H.} 2009.
\newblock The connectivity of boolean satisfiability: Computational and
  structural dichotomies.
\newblock {\em {SIAM} Journal on Computing\/}~{\em 38,\/}~6, 2330--2355.

\bibitem[\protect\citeauthoryear{Haas and Seyffarth}{Haas and
  Seyffarth}{2014}]{Haasey14}
{\sc Haas, R.} {\sc and} {\sc Seyffarth, K.} 2014.
\newblock The k-dominating graph.
\newblock {\em Graphs and Combinatorics\/}~{\em 30,\/}~3, 609--617.

\bibitem[\protect\citeauthoryear{Haddadan, Ito, Mouawad, Nishimura, Ono,
  Suzuki, and Tebbal}{Haddadan et~al\mbox{.}}{2016}]{haitmonionsute16}
{\sc Haddadan, A.}, {\sc Ito, T.}, {\sc Mouawad, A.~E.}, {\sc Nishimura, N.},
  {\sc Ono, H.}, {\sc Suzuki, A.}, {\sc and} {\sc Tebbal, Y.} 2016.
\newblock The complexity of dominating set reconfiguration.
\newblock {\em Theoretical Computer Science\/}~{\em 651}, 37--49.

\bibitem[\protect\citeauthoryear{Hearn and Demaine}{Hearn and
  Demaine}{2005}]{headem05}
{\sc Hearn, R.~A.} {\sc and} {\sc Demaine, E.~D.} 2005.
\newblock Pspace-completeness of sliding-block puzzles and other problems
  through the nondeterministic constraint logic model of computation.
\newblock {\em Theoretical Computer Science\/}~{\em 343,\/}~1-2, 72--96.

\bibitem[\protect\citeauthoryear{Hearn and Demaine}{Hearn and
  Demaine}{2009}]{Headem09}
{\sc Hearn, R.~A.} {\sc and} {\sc Demaine, E.~D.} 2009.
\newblock {\em Games, puzzles and computation}.
\newblock A {K} Peters.

\bibitem[\protect\citeauthoryear{Hirate, Banbara, Inoue, Lu, Nabeshima, Schaub,
  Soh, and Tamura}{Hirate et~al\mbox{.}}{2023}]{hibainlunascsota23}
{\sc Hirate, T.}, {\sc Banbara, M.}, {\sc Inoue, K.}, {\sc Lu, X.}, {\sc
  Nabeshima, H.}, {\sc Schaub, T.}, {\sc Soh, T.}, {\sc and} {\sc Tamura, N.}
  2023.
\newblock Hamiltonian cycle reconfiguration with answer set programming.
\newblock In {\em Proceedings of the 18th Edition of the European Conference on
  Logics in Artificial Intelligence (JELIA 2023)}, {S.~A. Gaggl}, {M.~V.
  Martinez}, {and} {M.~Ortiz}, Eds. Lecture Notes in Computer Science, vol.
  14281. Springer, 262--277.

\bibitem[\protect\citeauthoryear{Huynh}{Huynh}{2020}]{Huynh20}
{\sc Huynh, M.~K.} 2020.
\newblock Solving dominating set using answer set programming.
\newblock M.S.\ thesis, Heinrich Heine Universit\"{a}t d\"{u}sseldorf.

\bibitem[\protect\citeauthoryear{Inoue, Takano, Watanabe, Kawahara, Yoshinaka,
  Kishimoto, Tsuda, Minato, and Hayashi}{Inoue
  et~al\mbox{.}}{2014}]{itwkyktmh14}
{\sc Inoue, T.}, {\sc Takano, K.}, {\sc Watanabe, T.}, {\sc Kawahara, J.}, {\sc
  Yoshinaka, R.}, {\sc Kishimoto, A.}, {\sc Tsuda, K.}, {\sc Minato, S.}, {\sc
  and} {\sc Hayashi, Y.} 2014.
\newblock Distribution loss minimization with guaranteed error bound.
\newblock {\em {IEEE} Transactions Smart Grid\/}~{\em 5,\/}~1, 102--111.

\bibitem[\protect\citeauthoryear{Ito, Demaine, Harvey, Papadimitriou, Sideri,
  Uehara, and Uno}{Ito et~al\mbox{.}}{2011}]{itdehapasiueun11}
{\sc Ito, T.}, {\sc Demaine, E.~D.}, {\sc Harvey, N. J.~A.}, {\sc
  Papadimitriou, C.~H.}, {\sc Sideri, M.}, {\sc Uehara, R.}, {\sc and} {\sc
  Uno, Y.} 2011.
\newblock On the complexity of reconfiguration problems.
\newblock {\em Theoretical Computer Science\/}~{\em 412,\/}~12-14, 1054--1065.

\bibitem[\protect\citeauthoryear{Ito, Kawahara, Nakahata, Soh, Suzuki,
  Teruyama, and Toda}{Ito et~al\mbox{.}}{2023}]{itkanasosuteto23}
{\sc Ito, T.}, {\sc Kawahara, J.}, {\sc Nakahata, Y.}, {\sc Soh, T.}, {\sc
  Suzuki, A.}, {\sc Teruyama, J.}, {\sc and} {\sc Toda, T.} 2023.
\newblock {ZDD}-based algorithmic framework for solving shortest
  reconfiguration problems.
\newblock In {\em Proceedings of the 20th International Conference on the
  Integration of Constraint Programming, Artificial Intelligence, and
  Operations Research ({CPAIOR} 2023)}, {A.~A. Cir{\'{e}}}, Ed. Lecture Notes
  in Computer Science, vol. 13884. Springer, 167--183.

\bibitem[\protect\citeauthoryear{Ito, Ono, and Otachi}{Ito
  et~al\mbox{.}}{2015}]{itonot15}
{\sc Ito, T.}, {\sc Ono, H.}, {\sc and} {\sc Otachi, Y.} 2015.
\newblock Reconfiguration of cliques in a graph.
\newblock In {\em Proceedings of the 12th Annual Conference on Theory and
  Applications of Models of Computation (TAMC 2015)}, {R.~Jain}, {S.~Jain},
  {and} {F.~Stephan}, Eds. Lecture Notes in Computer Science, vol. 9076.
  Springer, 212--223.

\bibitem[\protect\citeauthoryear{Kaminski, Medvedev, and Milanic}{Kaminski
  et~al\mbox{.}}{2012}]{kamemi12}
{\sc Kaminski, M.}, {\sc Medvedev, P.}, {\sc and} {\sc Milanic, M.} 2012.
\newblock Complexity of independent set reconfigurability problems.
\newblock {\em Theoretical Computer Science\/}~{\em 439}, 9--15.

\bibitem[\protect\citeauthoryear{Knuth}{Knuth}{2011}]{knuth11}
{\sc Knuth, D.} 2011.
\newblock {\em The art of computer programming Vol.4A, Combinatorial
  algorithms, Part 1}.
\newblock Addison-Wesley.

\bibitem[\protect\citeauthoryear{Lifschitz}{Lifschitz}{2019}]{lifschitz19a}
{\sc Lifschitz, V.} 2019.
\newblock {\em Answer Set Programming}.
\newblock Springer-Verlag.

\bibitem[\protect\citeauthoryear{Mohar and Salas}{Mohar and
  Salas}{2009}]{mohsal09}
{\sc Mohar, B.} {\sc and} {\sc Salas, J.} 2009.
\newblock A new kempe invariant and the (non)-ergodicity of the
  wang–swendsen–kotecký algorithm.
\newblock {\em Journal of Physics A: Mathematical and Theoretical\/}~{\em
  42,\/}~22 (may), 225204.

\bibitem[\protect\citeauthoryear{Mouawad, Nishimura, Pathak, and Raman}{Mouawad
  et~al\mbox{.}}{2017}]{monipara17}
{\sc Mouawad, A.~E.}, {\sc Nishimura, N.}, {\sc Pathak, V.}, {\sc and} {\sc
  Raman, V.} 2017.
\newblock Shortest reconfiguration paths in the solution space of boolean
  formulas.
\newblock {\em SIAM Journal on Discrete Mathematics\/}~{\em 31,\/}~3,
  2185--2200.

\bibitem[\protect\citeauthoryear{Nishimura}{Nishimura}{2018}]{nishimura18}
{\sc Nishimura, N.} 2018.
\newblock Introduction to reconfiguration.
\newblock {\em Algorithms\/}~{\em 11,\/}~4, 52.

\bibitem[\protect\citeauthoryear{Soh, Tanjo, Okamoto, and Ito}{Soh
  et~al\mbox{.}}{2024}]{sotaokit24}
{\sc Soh, T.}, {\sc Tanjo, T.}, {\sc Okamoto, Y.}, {\sc and} {\sc Ito, T.}
  2024.
\newblock {CoRe} challenge 2022/2023: Empirical evaluations for independent set
  reconfiguration problems (extended abstract).
\newblock In {\em Proceedings of the Seventeenth International Symposium on
  Combinatorial Search ({SOCS} 2024)}, {A.~Felner} {and} {J.~Li}, Eds. {AAAI}
  Press, 285--286.

\bibitem[\protect\citeauthoryear{Suzuki, Mouawad, and Nishimura}{Suzuki
  et~al\mbox{.}}{2016}]{sumoni16}
{\sc Suzuki, A.}, {\sc Mouawad, A.~E.}, {\sc and} {\sc Nishimura, N.} 2016.
\newblock Reconfiguration of dominating sets.
\newblock {\em Journal of Combinatorial Optimization\/}~{\em 32,\/}~4,
  1182--1195.

\bibitem[\protect\citeauthoryear{Takaoka}{Takaoka}{2018}]{takaoka18}
{\sc Takaoka, A.} 2018.
\newblock Complexity of hamiltonian cycle reconfiguration.
\newblock {\em Algorithms\/}~{\em 11,\/}~9, 140.

\bibitem[\protect\citeauthoryear{Toda, TakehiroIto, Kawahara, Soh, Suzuki, and
  Teruyama}{Toda et~al\mbox{.}}{2023}]{tiksut23}
{\sc Toda, T.}, {\sc TakehiroIto}, {\sc Kawahara, J.}, {\sc Soh, T.}, {\sc
  Suzuki, A.}, {\sc and} {\sc Teruyama, J.} 2023.
\newblock Solving reconfiguration problems of first-order expressible
  properties of graph vertices with boolean satisfiability.
\newblock In {\em Proceedings of the IEEE 35th International Conference on
  Tools with Artificial Intelligence (ICTAI 2023)}. 294--302.

\bibitem[\protect\citeauthoryear{van~den Heuvel}{van~den
  Heuvel}{2013}]{heuvel13}
{\sc van~den Heuvel, J.} 2013.
\newblock The complexity of change.
\newblock In {\em Surveys in Combinatorics 2013}, {S.~R. Blackburn},
  {S.~Gerke}, {and} {M.~Wildon}, Eds. London Mathematical Society Lecture Note
  Series, vol. 409. Cambridge University Press, 127--160.

\bibitem[\protect\citeauthoryear{Yamada, Banbara, Inoue, and Schaub}{Yamada
  et~al\mbox{.}}{2023}]{yabainsc23}
{\sc Yamada, Y.}, {\sc Banbara, M.}, {\sc Inoue, K.}, {\sc and} {\sc Schaub,
  T.} 2023.
\newblock Recongo: Bounded combinatorial reconfiguration with answer set
  programming.
\newblock In {\em Proceedings of the 18th Edition of the European Conference on
  Logics in Artificial Intelligence (JELIA 2023)}, {S.~A. Gaggl}, {M.~V.
  Martinez}, {and} {M.~Ortiz}, Eds. Lecture Notes in Computer Science, vol.
  14281. Springer, 278--286.

\bibitem[\protect\citeauthoryear{Yamada, Banbara, Inoue, Schaub, and
  Uehara}{Yamada et~al\mbox{.}}{2024}]{ybisu24}
{\sc Yamada, Y.}, {\sc Banbara, M.}, {\sc Inoue, K.}, {\sc Schaub, T.}, {\sc
  and} {\sc Uehara, R.} 2024.
\newblock Combinatorial reconfiguration with answer set programming:
  Algorithms, encodings, and empirical analysis.
\newblock In {\em Proceedings of the 18th International Conference and
  Workshops on Algorithms and Computation (WALCOM 2024)}, {R.~Uehara},
  {K.~Yamanaka}, {and} {H.~Yen}, Eds. Lecture Notes in Computer Science, vol.
  14549. Springer, 242--256.

\end{thebibliography}

\end{document}